\documentclass{article}

\usepackage[preprint]{neurips_2026}

\usepackage{amsmath,amsfonts,bm}

\def\eqref#1{equation~\ref{#1}}

\def\1{\bm{1}}

\DeclareMathAlphabet{\mathsfit}{\encodingdefault}{\sfdefault}{m}{sl}
\SetMathAlphabet{\mathsfit}{bold}{\encodingdefault}{\sfdefault}{bx}{n}

\usepackage{hyperref}
\usepackage{url}

\usepackage{graphicx}
\usepackage{booktabs}
\usepackage{mathrsfs}
\usepackage{amsmath}
\usepackage{enumitem}
\usepackage{wrapfig}
\usepackage{array}
\usepackage{caption}
\usepackage{marvosym}
\usepackage{comment}
\usepackage{pifont}
\usepackage{tabularx}
\usepackage{makecell}
\usepackage{comment}
\usepackage{natbib}
\usepackage[breakable]{tcolorbox}
\usepackage{colortbl}
\definecolor{lightgray}{gray}{0.93}

\usepackage[utf8]{inputenc} 
\usepackage[T1]{fontenc}    
\usepackage{hyperref}       
\usepackage{url}            
\usepackage{booktabs}       
\usepackage{amsfonts}       
\usepackage{nicefrac}       
\usepackage{microtype}      
\usepackage{xcolor}         

\title{LLM as Graph Kernel: Rethinking Message Passing on Text-Rich Graphs}

\author{
    Ying Zhang$^{1}$\thanks{Work done during an internship at Ant Group.}~~~Hang Yu$^{1}$\thanks{Corresponding author.}~~~Haipeng Zhang$^{2}$\footnotemark[2]~~~Peng Di$^{1}$ \\
    ${}^1$Ant Group~~~${}^2$ShanghaiTech University\\
    \texttt{sgjzp.joyce@gmail.com}\\
    \texttt{\{hyu.hugo, dipeng.dp\}@antgroup.com}\\
    \texttt{haipengzhang@shanghaitech.edu.cn}
}

\begin{document}

\maketitle

\begin{abstract}

Text-rich graphs, which integrate complex structural dependencies with abundant textual information, are ubiquitous yet remain challenging for existing learning paradigms. 
Conventional methods and even LLM-hybrids compress rich text into static embeddings or summaries before structural reasoning, creating an information bottleneck and detaching updates from the raw content.
We argue that in text-rich graphs, the text is not merely a node attribute but the primary medium through which structural relationships are manifested.
We introduce \textbf{RAMP}, a \textbf{R}aw-text \textbf{A}nchored \textbf{M}essage \textbf{P}assing approach that moves beyond using LLMs as mere feature extractors and instead \textbf{recasts the LLM itself as a graph-native aggregation operator}.
RAMP exploits the text-rich nature of the graph via a novel dual-representation scheme: it anchors inference on each node’s raw text during each iteration while propagating dynamically optimized messages from neighbors. It further handles both discriminative and generative tasks under a single unified generative formulation.
\setcounter{footnote}{0}
Extensive experiments show that RAMP effectively bridges the gap between graph propagation and deep text reasoning, achieving competitive performance and offering new insights into the role of LLMs as graph kernels for general-purpose graph learning.\footnote{The code is available at \url{https://github.com/codefuse-ai/CodeFuse-RAMP}}

\end{abstract}

\vspace{-1.5ex}
\section{Introduction}
\label{sec:intro}
\vspace{-1ex}

Graph representation learning has emerged as a cornerstone for modeling real-world complex data, with applications in diverse domains such as social networks, knowledge graphs, and recommendation systems~\citep{jin2024large,zhang2024oag,wang2025graph}. Its impressive generalization ability across diverse graph structures largely stems from the message passing mechanism, which lies at the heart of modern Graph Neural Networks (GNNs)~\citep{ma2021deep,cai2023connection}. This mechanism is crucial for overcoming the \textit{structural bottleneck}—the inherent limitation of processing nodes independently—and is widely applied to text-rich graphs, where nodes or edges are associated with natural language~\citep{zhao2022learning,zhu2025llm}. \textbf{However, we observe a fundamental mismatch in current text-rich graph learning: while the graph is ``text-rich,'' the message-passing process is ``text-starved.''}
Indeed, the learning process on text-rich graphs typically involves two main stages: an initialization stage that transforms each node's raw text into a dense \textit{semantic} representation, and an aggregation stage that iteratively passes these representations (message) among neighbors to capture \textit{structural} information~\citep{jin2024large}.

While effective at mitigating the structural bottleneck, this ``encode-then-aggregate'' paradigm suffers from a pervasive semantic bottleneck characterized by irreversible information loss. Because traditional GNNs lack the capacity to ingest raw text during the update step, they are forced to ``freeze'' the semantics during initialization~\citep{zhao2022learning,jin2024large}. This creates a decoupling where structural updates occur in a vacuum, isolated from the nuanced linguistic context that defines the nodes.
During the \textbf{initialization stage}, raw, often lengthy, text is aggressively compressed into a single fixed-size vector by a text encoder~\citep{zhao2022learning,zhang2024hierarchical}, collapsing the semantic manifold and discarding fine-grained semantic details. This initial loss is catastrophic for text-rich graphs, where the solution to a task (e.g., node classification or graph QA) often depends on a specific phrase or relationship buried within the raw text that a static embedding cannot capture.
\begin{figure*}[!t]
\setlength{\belowcaptionskip}{-3pt}
    \centering
    \includegraphics[width=1\linewidth]{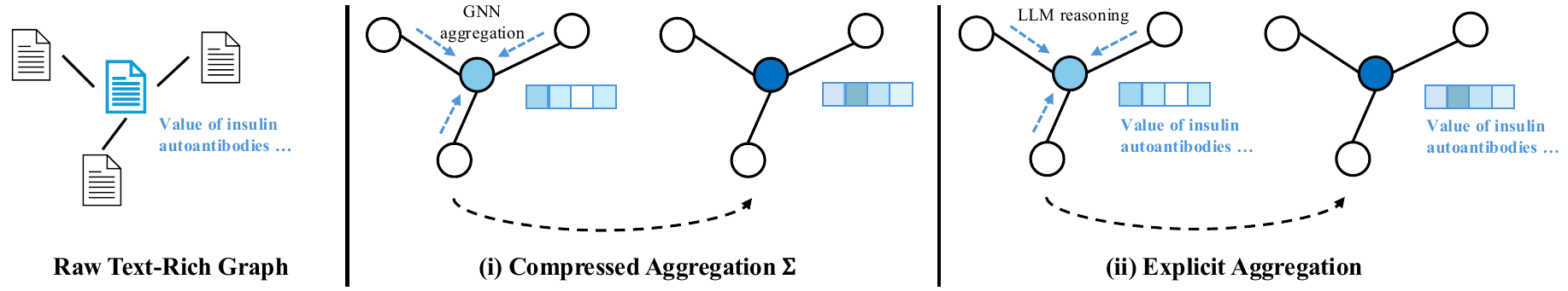}
    \caption{Illustration of message passing paradigms on text-rich graphs. Given a text-rich graph (left) where each node is associated with long textual content, (i) traditional GNNs compress node texts into a compact representation during aggregation, whereas (ii) explicit aggregation paradigm retains original text for LLM reasoning.}
    \label{fig:ramp_intro}
\end{figure*}

Recently, the advent of Large Language Models (LLMs) offers new avenues for tackling the semantic bottleneck~\citep{chen2024llaga,zhang2024hierarchical,zhu2025llm}. One radical approach is to linearize the graph or subgraph structure and its textual content into a single text sequence, feeding it directly to an LLM~\citep{mavromatis2024gnn}. While this approach preserves the integrity of the original text, sacrificing the explicit graph topology reintroduces the very structural bottleneck that message passing was designed to solve. Seeking a balance between structure and semantics, current approaches employ LLMs either by architecturally integrating external modules for graph learning~\citep{liu2024one,kong2024gofa,yang2024gl} or by behaviorally simulating message passing via hard/soft prompting~\citep{zhang2024hierarchical,zhu2025llm}. Despite this tighter integration, these methods still perform the aggregation of message on intermediate representations rather than the original raw text, leaving the semantic bottleneck essentially unaddressed.

Overcoming this requires a paradigm shift: \textbf{instead of merely using an LLM to process text for a GNN, we must empower the LLM to become the GNN}. To this end, we argue that a true LLM-native GNN must maintain a persistent, non-compressive link to a node's primary content while simultaneously performing structural aggregation as shown in Fig.~\ref{fig:ramp_intro}(ii). We propose RAMP, which recasts the message-passing update as an in-context reasoning task. The key insight of RAMP is the ``Semantic Anchor'': by inputting the target node's raw text at every layer, we ensure that the structural messages from neighbors are always interpreted through the lens of the original, uncompressed textual evidence. Unlike prior methods that choose between graph topology (GNNs) or semantic depth (LLMs), RAMP introduces an asymmetric dual-representation aggregation scheme. This allows the model to treat the target node as a high-fidelity semantic anchor (raw text) while treating its neighborhood as a compressed, optimizable memory space (summary tokens). Our contributions are: 

\begin{figure*}[!t]
\setlength{\belowcaptionskip}{-3pt}
    \centering
    \includegraphics[width=1\linewidth]{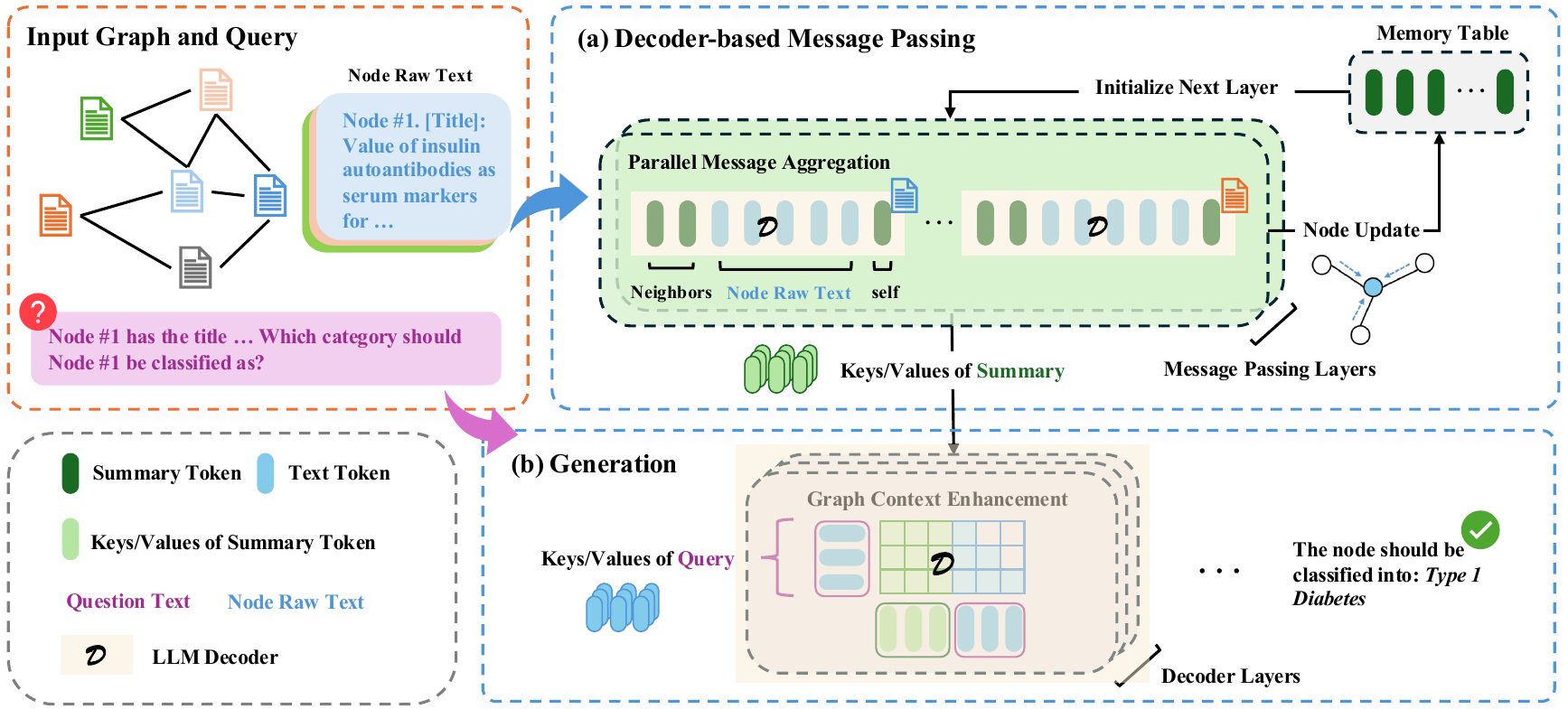}
    \caption{Architecture of RAMP. Given a text-rich graph and a query, we wrap node contents into a token sequence and perform parallel decoding on all nodes to obtain their summaries.
    Hidden states of summary tokens are stored in a memory table to initialize the next layer, realizing (a) layer-wise message-passing; the final decoder aggregates graph information for (b) the answer generation.}
    \label{fig:ramp_arch}
\end{figure*}

\begin{itemize}[leftmargin=*,itemsep=0pt,topsep=0pt,partopsep=0pt]
    \item We characterize the ``Semantic-Structural Decoupling'' problem in text-rich graphs, showing that current GNNs fail because they separate text encoding from structural propagation.
    
    \item We propose the ``Semantic Anchor'' mechanism, a novel message-passing design that enables an LLM decoder to function as a graph kernel by processing raw text and structural messages simultaneously.
    
    \item We demonstrate that by maintaining text fidelity at every layer, RAMP achieves a strong balance between structural awareness and deep semantic reasoning.
    
\end{itemize}
\section{Related Work}
\label{sec:related_work_main}

We briefly outline three lines of related work here; a comprehensive discussion is provided in App.~\ref{sec:related_work}.

\paragraph{Text-Rich Graph}
Current approaches predominantly follow an \textbf{encode-then-aggregate} pipeline: a text encoder first compresses each node's raw text into a fixed-size embedding, which is then propagated through a GNN~\citep{jin2024large,zhao2022learning}.
While computationally efficient, this two-stage process creates a semantic bottleneck: fine-grained textual cues are irreversibly lost before any structural reasoning begins.
Joint training schemes such as GLEM~\citep{zhao2022learning} and adapter-based methods~\citep{li2023graphadapter} alleviate but do not eliminate this bottleneck, as the aggregation stage still operates on compressed representations.

\paragraph{Message Passing in GNNs}
The message-passing paradigm~\citep{hamilton2017inductive,velivckovic2017graph} underpins modern GNNs.
A key insight from attention-based GNNs (e.g., GAT~\citep{velivckovic2017graph}) is the asymmetric query--key design: a node's own
representation actively probes its neighborhood rather than passively averaging neighbor features.
RAMP extends this principle to the LLM era by treating a node's \textbf{full raw text} as the query and its neighbors' compact summary
tokens as keys/values, enabling context-aware aggregation grounded in the original linguistic evidence.

\paragraph{Graphs in LLMs}
Recent efforts to unify LLMs with graph learning fall into three categories:
(i)~\textbf{graph serialization}~\citep{mavromatis2024gnn,ye2023language}, which linearizes the graph into text but sacrifices structural
nativity;
(ii)~\textbf{architectural hybrids} such as GOFA~\citep{kong2024gofa} and GL-Fusion~\citep{yang2024gl}, which inject GNN-like modules
into an LLM; and
(iii)~\textbf{behavioral simulation} such as PromptGFM~\citep{zhu2025llm} and HiCom~\citep{zhang2024hierarchical}, which prompt an LLM to
emulate GNN aggregation.
Despite their sophistication, these categories do not realize iterative, structure-aware message passing grounded in raw text at each propagation layer.
RAMP departs from these approaches by operationalizing the LLM decoder \textbf{itself} as the core message-passing operator, maintaining
a persistent link to each node's raw text at every propagation layer.
\vspace{-1ex}
\section{Method}
\vspace{-1ex}
\label{sec:method}

This section presents the RAMP framework.
We first revisit the conventional message passing in GNNs to pinpoint the semantic bottleneck.
We then argue that LLMs, particularly their inherent ability to reason over raw text, can serve as a \emph{graph kernel} that naturally resolves this challenge. 
Finally, we introduce RAMP as an LLM-native GNN counterpart with its design and training recipe. 

\vspace{-1ex}
\subsection{Revisiting GNNs and the Semantic-Structural Decoupling}
\vspace{-1ex}

The core of modern GNNs is the \textbf{message passing} paradigm, where each node updates its representation by aggregating information from its local neighborhood.
Following the general formulation that unifies models like MPNN~\citep{cai2023connection} and GAT~\citep{velivckovic2017graph}, a GNN layer is generally defined as:
\begin{equation}
\mathbf{h}_i^{(\ell+1)} = \textbf{Update}
\,\!\Big(
\mathbf{h}_{i}^{(\ell)}, {\textbf{Aggregate}
( \{
\textbf{Msg}(\mathbf{h}_{j}^{(\ell)}):
j \in \mathcal{N}(i)
\})
}
\Big),
\label{eq:gnn}
\end{equation}
where $\mathbf{h}_i^{(\ell)}$ is the state of node $i$ at layer $\ell$, and $\mathcal{N}(i)$ is its neighbor set.
This process involves three key functions:
\textbf{1. `Msg':} Transforms a neighbor's state $\mathbf{h}_j^{(\ell)}$ into a message.
\textbf{2. `Aggregate':} A permutation-invariant function (e.g., sum, mean, or attention) that combines messages from all neighbors into a single vector.
\textbf{3. `Update':} Fuses the aggregated message with the target node's own state $\mathbf{h}_i^{(\ell)}$ to produce the new state $\mathbf{h}_i^{(\ell+1)}$.
Often, this step includes a residual or skip connection to $\mathbf{h}_i^{(\ell)}$ to preserve the node's identity and stabilize training in deep models.
This formulation reveals a crucial \textbf{asymmetric treatment} of the target node versus its neighbors.
The target node’s state, $\mathbf{h}_i^{(\ell)}$, is preserved as a distinct \textbf{self-representation} that guides the aggregation. Taking GAT~\citep{velivckovic2017graph} as an example, $\mathbf{h}_i^{(\ell)}$ acts as a \textbf{query} to compute attention weights over its neighbors' states, which serve as \textbf{keys} and \textbf{values}.
This allows the model to learn neighbor relevance, enabling focused, context-aware updates. While powerful, this paradigm faces a fundamental \textbf{semantic bottleneck} when applied to text-rich graphs. The structural update (Update) happens iteratively, whereas the semantic content is typically ``frozen'' during a one-time initialization stage.
This premature compression of long, nuanced documents into a single vector inevitably discards critical semantic details, a problem well-documented in the literature~\citep{zhang2024hierarchical}.
Consequently, the GNN's message-passing layers are \textbf{``text-starved''}: they perform structural reasoning in a semantic vacuum, isolated from the raw text that defines the nodes. Any nuance not captured in the initial $\mathbf{h}_i^{(0)}$ is permanently lost to the structural propagation.
This begs the question: \textit{how can we build a GNN counterpart that performs message passing directly on raw text without sacrificing scalability?}

\vspace{-1ex}
\subsection{LLMs as Natural Graph Propagators}
\label{sec:motivation}
\vspace{-1ex}

We find a compelling answer in recent advancements in long-context LLMs. These models have faced a parallel challenge: the quadratic complexity of self-attention limits their context window. A highly successful solution has been \emph{context compression}, where the LLM itself is adapted to distill long sequences of text into a few compact, special-purpose tokens. These tokens, referred to as ``summary vectors''~\citep{chevalier2023adapting}, ``memory slots''~\citep{ge2023context}, or ``activation beacons''~\citep{zhang2024soaring}, function as learned, compact representations—or embeddings—of the original text. Crucially, models like ICAE~\citep{ge2023context} and AutoCompressor~\citep{chevalier2023adapting} have shown that an LLM can be trained via a simple auto-encoding objective to generate these summary tokens, from which the original text can be faithfully reconstructed. This demonstrates that an LLM is a powerful and natural tool for information-preserving compression. This principle of LLM-driven context compression perfectly fills the gap in text-rich GNNs.
We propose to move beyond ``text-starved'' propagation by recasting the LLM itself as the graph propagation mechanism. In our framework, the LLM’s self-attention mechanism is not just for language modeling; it becomes the engine for performing the GNN Aggregate and Update operations directly,
where the raw text remains central to every update.
Concretely, we make the following critical mapping:
(1) The target node's high-fidelity state $\mathbf{h}_i^{(\ell)}$ is instantiated as its \textbf{raw text, }$\mathbf{X}_i$.
(2) The messages from neighbors, $\textbf{Msg}(\mathbf{h}_j^{(\ell)})$, are instantiated as compact \textbf{summary tokens, }$\mathbf{S}_j^{(\ell)}$.
By leveraging an LLM decoder as the core computational unit, RAMP creates an elegant and powerful framework for scalable, structurally faithful learning on text-rich graphs.

\vspace{-1ex}
\subsection{RAMP: Raw-text Anchored Message Passing}
\label{sec:ramp_detail}

The RAMP framework shown in Fig.~\ref{fig:ramp_arch} consists of three main stages, mirroring the core workflow of GNNs: (1) initializing node representations, (2) performing multi-hop message passing, and (3) making a final prediction. By recasting the standard Transformer block as a graph-native reasoning operator, RAMP decomposes the message-passing process into a token-level reasoning task. For clarity, we use $\mathbf{h}_i^{(\ell)}$ to denote the state of node $i$ at layer $\ell$ in a classical GNN, and $\mathbf{S}_i^{(\ell)}$ (the summary tokens) for its counterpart in RAMP.

\vspace{0.5ex}
\noindent\textbf{Step 1: Semantic-Preserving Initialization via Ratio-Based Compression.}~While classical GNNs initialize each node with a single state $\mathbf{h}_i^{(0)}$, RAMP preserves the text-rich nature of the graph by creating a dual representation:
we maintain the raw text $\mathbf{X}_i$ as the primary semantic reference and generate a parallel sequence of summary tokens $\mathbf{S}_i^{(0)}$ to act as the initial structural message.
This step can be viewed as creating the initial ``message'' that each node is capable of sending.
Inspired by parallel-decoding techniques~\citep{zhang2024soaring,li2024focusllm}, we prompt an LLM decoder to perform this compression. Concretely, for each node $i$, we append corresponding summary-token placeholders into the input text $\mathbf{X}_i$, and let the decoder generate hidden states for these special tokens:
$
\mathbf{S}_i^{(0)} = [s_{i,1}^{(0)}, \ldots, s_{i,n}^{(0)}], \, n = \lceil \rho \cdot L_i \rceil,
$
where $\rho \leq 1$ is a predefined compression ratio. Unlike fixed-size embeddings, this dynamic allocation ensures that nodes with richer text receive a higher-dimensional state space, preventing the ``one-size-fits-all'' information loss typical of fixed-vector initialization.

\vspace{0.5ex}
\noindent\textbf{Step 2: Asymmetric Message Passing via Contextual Parallel Decoding.}~This step unifies the GNN's `Aggregate' and `Update' functions into a single, parallelizable decoder inference step (as shown in Fig.~\ref{fig:ramp_arch}(a)). At each layer $\ell$, RAMP performs ``Anchor-Based Reasoning'':
for a target node $i$, the decoder does not merely update an abstract vector; instead, it re-evaluates the entire neighborhood through the lens of the raw text $\mathbf{X}_i$ (the Semantic Anchor):
\begin{equation}
\mathbf{S}_i^{(\ell+1)} = \textbf{Decoder}\Big(\big[\,
\underbrace{\mathbf{S}_{j_1}^{(\ell)} \,\|\, \cdots \,\|\, \mathbf{S}_{j_m}^{(\ell)}}_{\text{neighbors}} 
\,\|\, 
\underbrace{\mathbf{X}_i}_{\text{raw text}} 
\, \\ \|\,
\underbrace{\mathbf{S}_i^{(\ell)}}_{\text{self}}
\,\big]
\Big) ,\, j_1, ..., j_m \in \mathcal{N}(i).
\label{eq: input_sequence}
\end{equation}
This design directly operationalizes our central thesis of the LLM as a graph-native operator. It implements a novel dual-representation aggregation scheme within an LLM:
\textbf{(a) Neighbor Messages:} The summary tokens of neighbors, $\{\mathbf{S}_j^{(\ell)}\}$, are concatenated to form the aggregated neighborhood information, corresponding to $\textbf{Aggregate}(\{\textbf{Msg}(\mathbf{h}_j^{(\ell)})\})$ and representing the ``structural memory'' of the graph.
\textbf{(b) Self-State (The Semantic Anchor)}: The target node's \textbf{raw text}, $\mathbf{X}_i$, represents its full, high-fidelity state. By feeding $\mathbf{X}_i$ into the decoder at every layer, we ensure that the structural messages from neighbors are always interpreted in the context of the original evidence. This prevents the ``semantic drift'' that occurs in deep GNNs, where the original node content often becomes diluted by neighborhood noise.
\textbf{(c) Self-State (Prompt/Skip):} The target node's own summary tokens from the previous layer, $\mathbf{S}_i^{(\ell)}$, are also included.
This serves a dual role. Functionally, it acts as a \textbf{skip connection}, ensuring that the model can carry forward its previously learned representation. Mechanistically, by placing it at the end of the input sequence—after the neighbor summaries and the node's own raw text—it serves as the direct prompt for the auto-regressive decoder. A decoder generates an output sequence in a single forward pass, conditioned on a preceding context or prompt. To generate the \emph{new} summary sequence $\mathbf{S}_i^{(\ell+1)}$, we use the \emph{current} summary tokens $\mathbf{S}_i^{(\ell)}$ as the generative prompt. The decoder is thus tasked with ``editing'' or ``rewriting'' its input summary based on the provided context.
This turns message passing into an in-context revision task, where the LLM refines a node's summary by dynamically grounding neighborhood information against its own persistent semantic anchor.
The decoder processes this structured input and outputs a new sequence of hidden states, which constitute the updated summary tokens $\mathbf{S}_i^{(\ell+1)}$. These states are then stored in a memory table $\mathcal{M}$ to serve as messages for the next layer of propagation.

\vspace{0.5ex}
\noindent\textbf{Step 3: Generative Formulation and Decoupled KV Cache Materialization.}~After $L$ rounds of message passing, RAMP uses the final summary token representations to perform downstream tasks. Unlike classical GNNs that require task-specific heads, RAMP unifies all tasks under a single \textbf{generative formulation} (see Fig.~\ref{fig:ramp_arch}(b)). Here, the role of the summary tokens shifts from being \emph{states-to-be-updated} to being \emph{context-to-be-reasoned-with}.
This is where the Key-Value (KV) cache becomes essential. The final layer summary tokens, $\mathbf{S}_i^{(L)}$, provide two complementary forms of information: \textbf{1. Hidden States:} The final representations stored in the memory table, which embody the propagated node state. \textbf{2. Key-Value (KV) Cache:} Derived from these hidden states, the KV pairs are the mechanism by which the decoder can efficiently attend to the full graph context during auto-regressive generation.
\textbf{This deliberate separation of state propagation and final generation is a key design choice to ensure scalability}. During the iterative message passing in Step 2, we only propagate the compact hidden states. This is for two reasons. First, it mirrors how classical GNNs propagate state vectors layer by layer, where the goal is to compute the next layer's representation, not generate a final, human-readable answer. Second, it is significantly more memory-efficient. Storing the full KV cache for every node at every intermediate GNN layer would be prohibitively expensive, especially for deep LLMs and large graphs. The KV cache is only materialized at the very end, providing the final decoder with the necessary context for rich, generative reasoning without the massive overhead during propagation. 
For a given task query $Q$ and a target answer $Y = (y_1, \dots, y_T)$, we feed the aggregated KV caches from the relevant nodes into the final decoder.
Here the same decoder is used for both message passing and generation, with shared parameters.
The model is then trained end-to-end to maximize the conditional probability:
\begin{equation}
\mathcal{L}_\text{gen} = - \sum_{t=1}^T \log P(y_t \mid y_{<t}, Q, \mathbf{K}, \mathbf{V}),
\label{eq: generative_loss}
\end{equation}
where $(\mathbf{K}, \mathbf{V})$ are the final key-value caches. For node-level tasks, only the target node's KV cache is used. For graph-level tasks, the KV caches of all relevant nodes are provided, analogous to graph pooling in classical GNNs~\citep{ma2021deep}.

\newcommand{\tablesize}{\fontsize{9.5pt}{11.5pt}\selectfont} 

\begin{table*}[!thbp]
\setlength{\abovecaptionskip}{2pt}
\setlength{\belowcaptionskip}{-3pt}
\setlength{\extrarowheight}{4pt}
\setlength{\belowrulesep}{0.1ex} 
\setlength{\aboverulesep}{0.2ex} 
\centering
\caption{Connection between classical GNN message passing and the proposed RAMP.}
\tablesize
\begin{tabular}{ll}
\toprule
\textbf{GNN operation} & \textbf{RAMP counterpart} \\
\midrule
Node representation $\mathbf{h}_i^{(\ell)}$ 
& Summary token sequence $\mathbf{S}_i^{(\ell)} = [s_{i,1}^{(\ell)}, \dots, s_{i,n}^{(\ell)}]$ \\

Neighbor message $\{\mathbf{h}_j^{(\ell)} : j \in \mathcal{N}(i)\}$ 
& Concatenation of neighbor tokens $\mathbf{S}_{j}^{(\ell)}$ for $j \in \mathcal{N}(i)$ \\

Aggregation function $\text{GNN}(\mathbf{h}_i^{(\ell)}, \{\mathbf{h}_j^{(\ell)}\})$ 
& LLM reasons over $[\mathbf{S}_{j_1}^{(\ell)} \| \cdots \| \mathbf{S}_{j_m}^{(\ell)} \| \mathbf{X}_i \| \mathbf{S}_i^{(\ell)}]$ \\

Layer-wise propagation 
& Passing semantic messages via summary tokens $\mathbf{S}_i^{*}$ \\
\bottomrule
\end{tabular}
\label{tab:ramp_vs_gnn}
\end{table*}

\noindent\textbf{Discussion.}~Our proposed RAMP architecture is not merely an LLM-GNN hybrid; it represents an LLM-native
generalization of the classical GNN message-passing paradigm for text-rich graphs, as summarized in Tab.~\ref{tab:ramp_vs_gnn}.
A key advantage of RAMP is its \textbf{fidelity-preserving message passing}.
It achieves this by using a node's full, raw text ($\mathbf{X}_i$) to guide the attention over its neighbors' summary tokens $\mathbf{S}_j$, avoiding the irreversible semantic distortion inherent in traditional pipelines that operate solely on compressed embeddings $\mathbf{h}_j$.
This also replaces passive aggregation (e.g., mean or sum) with the LLM decoder’s own causal attention mechanism, turning aggregation into an active inference task over the structured input.

\vspace{-1ex}
\subsection{Training Recipe}
\label{sec:training_recipe}
\vspace{-0.5ex}

RAMP is trained in two consecutive stages: large-scale pre-training and task-specific fine-tuning.

\vspace{-1ex}
\paragraph{Pre-training.} Pre-training aims to teach the model two fundamental skills: text compression and message passing. We use two reconstruction-based tasks: \textbf{1. Self-reconstruction (Internal Fidelity):} The model must reconstruct a node's raw text $\mathbf{X}_i$ given only its own summary tokens $\mathbf{S}_i$. This trains the compression mechanism to be information-preserving, akin to the objective in FocusLLM~\citep{li2024focusllm}. \textbf{2.~Neighbor-reconstruction (Communicability):} The model must reconstruct $\mathbf{X}_i$ given its neighbors' summary tokens, $\{\mathbf{S}_j : j \in \mathcal{N}(i)\}$. This supervises the message-passing ability by encouraging the summary tokens to serve as interpretable messages that allow neighbors to infer content.
At each training step, we uniformly sample one of the two tasks and apply it to a randomly chosen target node $i$.
During pre-training, we update only the lightweight attention layers for the summary tokens, following efficient long-context fine-tuning practices~\citep{zhang2024soaring,li2024focusllm}. Further details on the pre-training setup are provided in App.~\ref{app:pretraining_setting}.

\vspace{-1ex}
\paragraph{Fine-tuning.} After pre-training, RAMP is adapted to downstream tasks using the same generative formulation in \eqref{eq: generative_loss} and fine-tuned efficiently with LoRA~\citep{hu2022lora}. A task-specific query (e.g., ``Classify this node'') is provided, and the model is trained to generate the desired output (e.g., ``The category is Machine Learning''). This unified framework allows RAMP to seamlessly handle both discriminative and generative graph learning tasks.

\vspace{-1ex}
\section{Experiment}
\label{sec:exp}
\vspace{-1ex}

This section empirically validates the core design principles of RAMP. Our experimental program is designed to answer three fundamental questions in sequence: (1) Does our core mechanism—the Semantic Anchor and dual-representation scheme—effectively preserve semantic information while enabling message passing? (2) Does this fidelity-preserving propagation translate into superior performance on downstream tasks that demand both deep textual understanding and structural reasoning? (3) Is our approach computationally practical, overcoming the scalability challenges that plague naive LLM-based graph methods? By addressing these questions, we demonstrate that RAMP successfully operates as an LLM-native graph learning architecture.

\vspace{-1ex}
\subsection{RAMP Pre-training: Validating Semantic Fidelity and Message Communicability}
\vspace{-0.5ex}
\label{ssec:ppl_exp}

\begin{wraptable}[13]{R}{0.5\linewidth}
\vspace{-12pt}
\small
\setlength{\tabcolsep}{2.0pt}
\begin{minipage}[t]{1\linewidth}
\caption{Evaluation for pre-trained RAMP. Qwen-2.5-7B-Instruct (21.72$^*$) is the backbone reference; its value is the raw language modeling perplexity obtained by directly continuing node text.}
\small
\centering
\begin{tabular*}{0.96\linewidth}{@{}lcc@{}}
\toprule
             & Perplexity$_\textit{self}$ $\downarrow$ & Perplexity$_\textit{nbr}$ $\downarrow$ \\ \midrule
Qwen-2.5-7B-Instruct   &    21.72*      &   -            \\ 
\midrule
RAMP$_{mp=1, \rho=0.05}$ &  11.36   &  20.68  \\ 
RAMP$_{mp=1, \rho=0.1}$ &  9.17   &  20.87  \\
RAMP$_{mp=2, \rho=0.1}$ & 8.93    & 22.97 \\
\midrule
RAMP$_{mp=2,compact}$ & 25.57    & 25.61 \\
\bottomrule
\end{tabular*}
\label{tab:exp_ppl}
\end{minipage}
\end{wraptable}


To verify that RAMP’s dual-representation scheme preserves information better than traditional static embeddings, we evaluate the quality of the summary tokens generated during pre-training. We pre-train RAMP on Qwen-2.5-7B-Instruct~\citep{qwen2.5}, following the training recipe in Sec.~\ref{sec:training_recipe}, and then measure its reconstruction perplexity (PPL) on the unseen Cora dataset. This test directly probes whether the ``Semantic-Structural Decoupling'' is resolved by assessing how much ``textual nuance'' is retained after compression.

Here we investigate two reconstruction modes: self-reconstruction (\emph{self},  internal fidelity), where a node’s text is recovered from the key–values of its own summary tokens, and neighbor-reconstruction (\emph{nbr}, communicability), where the text is generated based on the summary tokens of its neighbors.

To isolate the impact of our ``Semantic Anchor,'' we include a \emph{``compact''} baseline that performs message passing solely on summary tokens without re-accessing raw text.
We also vary message-passing rounds ($mp$) and compression ratios ($\rho$) to examine their effect on reconstruction quality.
The results in Tab.~\ref{tab:exp_ppl} validate the core mechanics of RAMP. First, the low PPL in self-reconstruction (e.g., 9.17 with $\rho=0.1$ vs. 21.72 for the raw backbone) confirms that RAMP’s summary tokens achieve high semantic fidelity, effectively compressing raw text into summary tokens with minimal information loss.
Second, the strong neighbor-reconstruction performance shows that our messages are ``globally interpretable'' across the graph, allowing a node to recover itself from neighbors' messages. Crucially, the ``compact'' variant’s failure (PPL >25) validates that without raw-text anchoring, even an LLM-based aggregator suffers from the semantic bottleneck common in traditional GNNs.
Moreover, this variant also shows consistent degradation on downstream node classification (see App.~\ref{ssec:exp_compact_downstream}), complementing the reconstruction-PPL evidence with task-level results.

Moreover, we observe an insightful trade-off with deeper propagation: while an additional message-passing round ($mp\!=\!2$) improves self-reconstruction, neighbor-reconstruction perplexity slightly increases. This suggests summary tokens evolve from pure compression into more abstract, context-aware representations. Although less suited for verbatim neighbor reconstruction, these 2-hop summaries are more powerful for high-level downstream tasks like classification (as shown in Sec.~\ref{ssec:struc_exp}), indicating a shift from simply summarizing content to learning a graph-aware function.


\vspace{-1ex}
\subsection{Unified Semantic and Structural Reasoning in Downstream Tasks}
\label{ssec:struc_exp}
\vspace{-0.5ex}

We now demonstrate how RAMP's core principle of unified reasoning translates into superior performance and robust structural awareness. Our primary evaluation benchmarks RAMP against representative baselines on \textbf{foundational text-rich graph tasks}. To further probe the model's genuine understanding of topology, we then subject it to a series of \textbf{structural perturbations}, including neighbor-order permutation and connectivity permutation.
Besides, an ablation study of the model's performance under different message-passing rounds is detailed in App~\ref{ssec:exp_mp_rounds}.

\vspace{-1ex}
\subsubsection{Foundational Graph Tasks}
\label{sssec:exp_graph_tasks}

\paragraph{Node Classification.} We evaluate RAMP on standard node classification benchmarks, comparing against (1) GNNs (GCN, GAT, GraphSAGE), (2) text-only PLMs (BERT, RoBERTa, Qwen-2.5-7B-Instruct), (3) recent GNN-LLM Integration methods (GOFA, OFA, LLaGA, and PromptGFM). More details are available in App.~\ref{sec:baselines}.

\begin{table*}[!t]
\setlength{\abovecaptionskip}{2pt}
\setlength{\belowcaptionskip}{-3pt}
\centering
\caption{Performance comparison of different methods on node classification tasks (Accuracy \% $\uparrow$). All baselines are evaluated under our unified setting with identical data splits. We highlight the best results in \textbf{bold} and the second best with \underline{underline}.}
\small
\begin{tabular}{l|cccccc}
\toprule
Method & Cora & Citeseer & PubMed & History & Photo & Arxiv \\
\midrule
\midrule
MLP & 69.00 & 59.35 & 75.55 & 77.71 & 47.56 & 57.23 \\
GCN~\citep{zhang2019graph} & 82.29 & 70.16 & 81.92 & 80.55 & 70.14 & 66.52 \\
GAT~\citep{velivckovic2017graph} & 83.03 & 71.29 & 80.78 & 78.93 & 66.18 & 67.76 \\
GraphSAGE~\citep{hamilton2017inductive} & 83.21 & 70.81 & 82.94 & 80.72 & 73.82 & 67.97 \\
\midrule
BERT~\citep{devlin2019bert} & 80.99 & 71.93 & \underline{91.75} & 80.94 & 58.34 & 65.64 \\
RoBERTa~\citep{liu2019roberta} & 76.93 & 70.48 & 91.37 & 79.42 & 57.27 & 66.72 \\
Qwen-2.5-7B-Instruct~\citep{qwen2.5} & 82.65 & \underline{74.35} & 90.79 & \underline{84.17} & 75.12 & \underline{72.99} \\
\midrule
GOFA~\citep{kong2024gofa} & 70.22 & 71.50 & 84.81 & 73.20 & 65.87 & 59.11 \\
OFA~\citep{liu2024one} & 81.73 & 74.19 & 86.91 & 81.75 & \textbf{77.23} & 64.93 \\
LLaGA~\citep{chen2024llaga} & 82.28 & 73.54 & 83.89 & 82.54 & 75.15 & 70.55 \\
PromptGFM (Qwen)~\citep{zhu2025llm} & \underline{83.95} & \underline{74.35} & 86.36 & 83.73 & 75.41 & 67.64 \\
\midrule
 \rowcolor{lightgray} \textbf{RAMP-7B} & \textbf{84.87} & \textbf{74.83} & \textbf{93.68} & \textbf{85.09} & \underline{76.21} & \textbf{75.38} \\
\bottomrule
\end{tabular}
\label{tab:exp_node_classification}
\vspace{-4ex}
\end{table*}


As shown in Tab.~\ref{tab:exp_node_classification}, RAMP-7B consistently outperforms both traditional GNNs and text-only PLM baselines across all datasets. This dual advantage stems directly from its unique architecture, which successfully unifies the strengths of both paradigms. \textbf{Against GNNs} like GraphSAGE, RAMP's superiority comes from its preservation of \textbf{Semantic Fidelity}; traditional GNNs suffer an information bottleneck by pre-compressing rich text into fixed embeddings, whereas RAMP’s message passing layers reason over the target node’s full raw text to capture nuanced semantics. Simultaneously, \textbf{against PLMs} like its own backbone (Qwen-2.5-7B-Instruct), RAMP's gains demonstrate the value of \textbf{Structural Integrity}. While the backbone LLM only processes isolated node texts, RAMP's message passing allows it to integrate rich contextual information from the multi-hop neighborhood for more informed predictions. By unifying deep text reasoning with graph propagation, RAMP thus overcomes the limitations of both prior approaches. Crucially, the architectural unification also allows RAMP to consistently surpass other hybrid methods like OFA and LLaGA across most benchmarks, achieving leads up to 6.77\% (on PubMed) and 4.83\% (on Arxiv).
This validates our raw-text anchored message-passing scheme, which maintains a persistent link to the semantic source at every layer, a feature absent in other integration models that aggregate already-compressed representations.

We also compare against two more recent baselines: GOFA~\citep{kong2024gofa}, a generative graph foundation model, and PromptGFM~\citep{zhu2025llm}, which prompts an LLM to emulate GNN aggregation.
To ensure a fair comparison under comparable computational budgets, we re-implement PromptGFM using the same Qwen-2.5-7B-Instruct backbone as RAMP (denoted PromptGFM (Qwen)), replacing its original GPT-4o dependency.
As shown in Tab.~\ref{tab:exp_node_classification}, RAMP-7B consistently outperforms both methods across nearly all datasets, with particularly large margins on PubMed (+8.87\% over GOFA, +7.32\% over PromptGFM (Qwen)) and Arxiv (+16.27\% over GOFA, +7.74\% over PromptGFM (Qwen)).
These results demonstrate that RAMP's raw-text anchored message-passing scheme yields consistent advantages over recent strong baselines across diverse datasets.

\begin{wraptable}[11]{L}{0.37\linewidth}
\vspace{-12pt}
\small
\setlength{\tabcolsep}{4.0pt}
\begin{minipage}[t]{1\linewidth}
\centering
\caption{Evaluation of RAMP on GraphQA tasks.}
\begin{tabular}{l|c} 
\toprule
Method & ExplaGraphs \\
\midrule
Qwen-2.5-7B-Instruct & 93.50 \\
\midrule
GNP~\citep{tian2024graph} & 87.04\\
G-Retriever~\citep{he2024g} & 87.05 \\
GRAG~\citep{Hu2024GRAGGR} & 88.05 \\
Align-GRAG~\cite{xu2025align} & 89.92 \\
\rowcolor{lightgray} \textbf{RAMP-7B} & \textbf{93.86} \\
\bottomrule
\end{tabular}
\label{tab:exp_graphqa}
\end{minipage}
\end{wraptable}


\vspace{-1ex}
\paragraph{GraphQA.} Beyond node classification, we evaluate RAMP on the GraphQA task, where the graph offers additional contextual information for answering questions. Tab.~\ref{tab:exp_graphqa} compares RAMP with its backbone, Qwen-2.5-7B-Instruct, and GNN-based baselines (including GNP, G-Retriever, GRAG, and Align-GRAG).

As shown in Tab.~\ref{tab:exp_graphqa}, RAMP-7B achieves 93.86\%, surpassing all baselines including its own backbone (93.50\%). The 0.36\% gain over the backbone may appear modest in absolute terms; however, this must be contextualized against the strength of the baseline. The backbone is not an off-the-shelf model but a heavily fine-tuned variant using the same LoRA configuration (rank=64) as RAMP (see App.~\ref{app:finetuing_setting}).
To provide a fuller picture, we note that the original, un-tuned Qwen-2.5-7B-Instruct achieves only 45.84\% in a zero-shot setting.
The massive gain from 45.84\% to 93.50\% is attributable to task-specific fine-tuning alone, while RAMP's further improvement represents a 5.5\% reduction in remaining error—a non-trivial step that the text-only model cannot achieve without graph structure.


\vspace{-1ex}
\subsubsection{Permutation Invariance and Structure Sensitivity}
\vspace{-0.5ex}
\label{ssec:exp_shuffling}

A true graph-aware model must exhibit two fundamental properties: (i) \textbf{invariance to the arbitrary order of neighbors}, and (ii) \textbf{sensitivity to the actual graph structure}. We design two shuffling tests to verify that RAMP has learned these properties. Correspondingly, the experimental settings are: \textbf{(1) Neighbor-Order Shuffle (Test for Invariance).} For each node in the graph, we randomly permute the order of its one-hop neighbors before feeding them into the model. This tests whether RAMP learns a permutation-invariant aggregation function, a core property of GNNs. \textbf{(2) Cross-Node Shuffle (Test for Sensitivity).} For each graph sample, we randomly swap the neighbor sets between different nodes. Each node is now connected to a random set of nodes from the same graph, breaking the original topology. This evaluates whether RAMP exploits meaningful signals from the graph structure or is merely a ``bag-of-neighbors'' model.

\begin{table}[!htbp]
\setlength{\abovecaptionskip}{2pt}
\setlength{\belowcaptionskip}{-3pt}
\centering
\caption{Impact of Shuffling (Accuracy \% $\uparrow$). We report the mean and standard deviation over three shuffling runs.}
\small
\begin{tabular}{c|cc}
\toprule
Setting & Cora & Citeseer \\
\midrule
RAMP (No Shuffle) & 84.87 & 74.83 \\
\midrule
Neighbor-Order Shuffle & 84.81 $\pm$ 0.08 & 74.73 $\pm$ 0.01 \\
Cross-Node Shuffle & 84.32 $\pm$ 0.14 & 74.41 $\pm$ 0.01 \\
\bottomrule
\end{tabular}
\label{tab:exp_ramp_node_shuffle}
\vspace{-2ex}
\end{table}


The results in Tab.~\ref{tab:exp_ramp_node_shuffle} provide strong evidence of RAMP’s structural integrity. First, performance remains stable when the order of neighbors is shuffled (Neighbor-Order Shuffle), with a negligible drop in accuracy and low variance. This confirms that RAMP learns a \textbf{permutation-invariant} aggregation function, behaving like a classical GNN. Second, performance degrades noticeably when the graph’s topology is broken (Cross-Node Shuffle), demonstrating that RAMP is not a mere “bag-of-neighbors” model but derives meaningful signals from the graph’s specific connectivity. This interpretation is further supported by a more radical ablation in App.~\ref{ssec:exp_cross_subgraph_shuffle}, where replacing neighbors with irrelevant cross-sample subgraphs causes performance to fall below the text-only backbone.

Collectively, the strong performance on downstream graph tasks (Sec.~\ref{sssec:exp_graph_tasks}), the consistent gains from deeper propagation (Sec.~\ref{ssec:exp_mp_rounds}), and the principled behavior in shuffle tests (Sec.~\ref{ssec:exp_shuffling}) provide comprehensive evidence that RAMP successfully maintains structural integrity, operating as a true graph-aware learning architecture.

\vspace{-1ex}
\subsection{Scalability of RAMP with Graph Size}
\vspace{-1.5ex}


\begin{wrapfigure}{R}{0.65\linewidth}
\vspace{-2ex}
\setlength{\abovecaptionskip}{2pt}
\setlength{\belowcaptionskip}{-3pt}
    \centering
    \includegraphics[width=\linewidth]{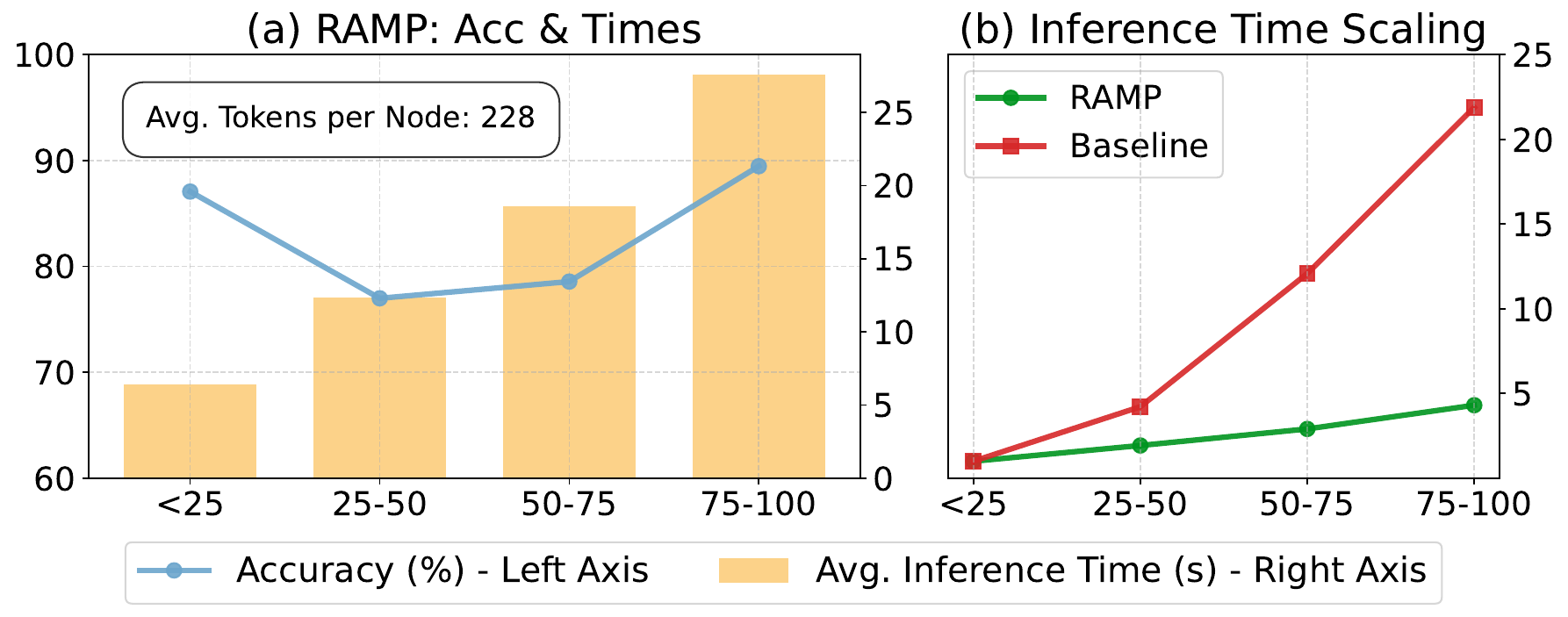}
    \caption{Scalability on Cora. (a) Accuracy and absolute inference time for RAMP. (b) Inference time scaling for each method, independently normalized to its performance in the smallest bucket (i.e., bucket $<25$).}
    \label{fig:graph_size_scale}
    \vspace{-3ex}
\end{wrapfigure}


Finally, we examine whether RAMP can scale effectively with increasing graph size.
We group subgraph samples into different size buckets (see App.~\ref{app:finetuing_setting} for subgraph construction) and report the corresponding node classification accuracy and inference time on Cora in Fig.~\ref{fig:graph_size_scale}.

Fig.~\ref{fig:graph_size_scale}(a) shows that RAMP-7B maintains stable accuracy on Cora even as subgraphs scale to nearly 100 nodes (see App.~\ref{ssec:scala_on_pubmed} for additional analysis on PubMed). To evaluate its efficiency, we compare RAMP against a Graph-to-Text baseline~\citep{fatemi2023talk}. While the 7B-size baseline triggers Out-of-Memory (OOM) due to $O(N^2)$ complexity, its 0.5B version reveals a sharp scaling contrast in Fig.~\ref{fig:graph_size_scale}(b). When normalized to their respective time on the smallest bucket, RAMP latency grows near-linearly by less than $5\times$, whereas the 0.5B baseline surges by $22\times$.
This disparity confirms that by assigning nodes to individual decoders and propagating compact summaries, RAMP enables raw-text anchored message passing without context-length explosion, highlighting its potential practicality and scalability for real-world graph applications.

\vspace{-1ex}
\section{Conclusion}
\label{sec:conclusion}
\vspace{-2ex}

We introduced RAMP, which redefines the LLM as a graph kernel that treats text as a dynamic medium for graph reasoning.
Its raw-text anchoring message-passing avoids prior information bottlenecks by directly integrating structural propagation with deep text understanding at each step.
Interestingly, intermediate message-passing rounds also yield meaningful generations (see App.~\ref{app:generation_across_layer}), which supports interpreting RAMP as a form of multi-agent communication where decoders function as agents~\citep{zou2025latent}. \paragraph{Limitations.} RAMP trades inference speed for semantic fidelity, and is currently validated on homogeneous graphs with well-established benchmarks. We provide a detailed discussion of limitations—including computational cost, potential data contamination, graph heterogeneity, and hyperparameter sensitivity in App.~\ref{app:limitations}.


\bibliographystyle{unsrt}
\bibliography{reference}

\clearpage
\appendix

\section*{Appendix}
\section{Ethical Considerations}
\label{app:ethical}
This work exclusively employs publicly available benchmark datasets, which involve no human subjects or sensitive data. We foresee no resulting ethical concerns or negative societal impact.


\section{Limitations}
\label{app:limitations}
We identify several limitations that suggest directions for future work.
\textbf{(1)~Computational cost.}
By design, RAMP trades inference speed for semantic fidelity.
Although its latency scales linearly with graph size (unlike the quadratic growth of Graph-to-Text methods), it remains substantially slower than classical GNNs.
Future work can leverage LLM acceleration techniques (e.g., quantization, speculative decoding) to narrow this gap.
\textbf{(2)~Potential data contamination.}
Our evaluation uses well-established benchmarks (e.g., Cora, PubMed) that may overlap with the pre-training corpora of the backbone LLM.
We mitigate this concern in two ways: (i)~RAMP's own pre-training stage uses only the MAPLE dataset from domains (Economics, Mathematics, Geology) disjoint from the downstream benchmarks, and (ii)~our primary comparison is against the \emph{same} fine-tuned backbone LLM, meaning any contamination advantage is shared equally by our baseline.
Nonetheless, evaluation on newly curated or contamination-controlled benchmarks would further strengthen the conclusions.
\textbf{(3)~Graph heterogeneity.}
All experiments are conducted on homogeneous graphs with uniform node and edge types.
Extending RAMP to heterogeneous or multi-relational graphs—where type-aware summary tokens or meta-path-guided aggregation may be required—is a significant and promising research direction.
\textbf{(4)~Hyperparameter sensitivity.}
While we provide empirical guidance for the compression ratio $\rho$ and message-passing depth (see App.~\ref{ssec:exp_compression_ratio} and~\ref{ssec:exp_mp_rounds}), a theoretical characterization of their optimal values remains open.

\section{Implementation Details}
\label{app:implementation}

We provide detailed descriptions of our implementation in this section. Specifically, we divide it into three parts: (1) settings for pre-training, where we describe the backbone model, optimization strategy, and training environment; (2) settings for downstream tasks, including fine-tuning configurations and prompt templates; and (3) training data construction, covering how we prepare both pre-training corpora and task-specific benchmarks.

\subsection{Settings for Pre-training}
\label{app:pretraining_setting}

We adopt Qwen-2.5-7B-Instruct~\citep{qwen2.5} as the backbone and follow the practices of FocusLLM~\citep{li2024focusllm} and Activation Beacon~\citep{zhang2024soaring} by freezing all backbone parameters and fully fine-tuning the additional attention layer.
Training is conducted on 32 NVIDIA A100 GPUs for one epoch over the entire pre-training corpora (see App.~\ref{app:training_data} for data construction). 
During pre-training, we adopt the optimization hyperparameters recommended in~\citep{li2024focusllm}, using BF16 mixed precision.
To reduce sensitivity to arbitrary neighbor ordering, we randomly permute each node’s neighbor order when constructing decoder inputs during pre-training, which encourages empirical permutation robustness.

\subsection{Settings for Downstream Tasks}
\label{app:finetuing_setting}

For fine-tuning on node-level tasks, we adopt the standard practice of constructing subgraph inputs~\citep{kong2024gofa}. 
Specifically, for a given target node, we expand its neighborhood to form an ego-graph, which is then fed into the model.
Furthermore, to provide task-specific context, the query itself is introduced as a special `prompt node' within this subgraph, which is connected to all other nodes.
To control input length, we set a maximum subgraph size and randomly subsample neighbors when the expansion exceeds this threshold.
Unless otherwise specified, all downstream experiments are conducted with the RAMP model pre-trained on Qwen-2.5-7B-Instruct, 
using two message-passing rounds and a compression ratio of $\rho\!=\!0.1$. 

During fine-tuning, we apply LoRA with rank~$\!=\!64$, $\alpha\!=\!64$, and dropout~$\!=\!0.1$.
We use early stopping with a patience of 3 epochs, based on validation accuracy.
The prompt templates we use for fine-tuning are displayed in Sec.~\ref{sec:prompt_template}.
All experiments are conducted on 16 NVIDIA A100 GPUs with a fixed random seed, producing deterministic results on identical hardware; we therefore report single-run numbers.

\subsection{Training Data Construction}
\label{app:training_data}

For pre-training, we construct corpora from the graph format of the MAPLE dataset~\citep{zhang2023effect}, a large-scale dataset for scientific literature tagging.
To ensure domain diversity while avoiding data leakage into downstream evaluations, we sample subgraphs from the following three domains: \textbf{Economics}, \textbf{Mathematics}, and \textbf{Geology}.
In total, we generate over \textbf{80,000} subgraph samples.
Each subgraph is obtained by randomly selecting a central node and then expanding along its edges to form a small ego-graph.
The subgraph size is uniformly sampled between 20 and 80 nodes, so that the model is exposed to graphs of varying scales while maintaining training efficiency.

Each node retains its raw textual content (i.e., paper title and abstract) as well as connectivity.
As described in Sec.~\ref{sec:training_recipe}, two reconstruction tasks are adopted during pre-training: 
(1) \textbf{self-reconstruction}, which performs reconstruction through the node's own summary tokens, and 
(2) \textbf{neighbor-reconstruction}, which conditions on the summary tokens of its neighbors when available, defaulting to self-reconstruction otherwise.

Fine-tuning is performed on task-specific benchmark datasets (e.g., Cora, Citeseer, PubMed), which are described in App.~\ref{app:dataset}.
\section{Dataset}
\label{app:dataset}

We evaluate RAMP on multiple benchmark datasets spanning different domains and tasks. Specifically, we consider (a) six node classification datasets that vary in domain and graph density, and (b) a GraphQA dataset for generative reasoning with structured evidence.

\begin{table*}[!thbp]
\centering
\setlength\tabcolsep{3pt}
\caption{Dataset Statistics. ``Avg. D'' means the average node degree. We additionally provide the domain and the composition of node text.}
\begin{tabular}{lrrccll}
\toprule
Dataset & \#Nodes & \#Edges & \#Classes & Avg. D & Domain & Node Text \\
\midrule
Cora        & 2,708 & 10,858 & 7 & 4.01  & Computer Science & Paper titles and abstracts \\
Citeseer       & 3,327 & 9,464 & 6 & 2.84  & Computer Science & Paper titles and abstracts \\
PubMed    & 19,717 & 88,648 & 3 & 4.49  & Biomedical   & Paper titles and abstracts \\
History  & 41,551 & 503,180 & 12 & 12.11  & E-commerce   &  Item titles and descriptions \\
Photo      & 48,362 & 873,782 & 12 & 18.07 & E-commerce  &  Item titles and reviews \\
Arxiv      & 169,343 & 1,166,243 & 40 & 6.89 & Computer Science  &  Paper titles and abstracts \\
\bottomrule
\end{tabular}
\label{tab:exp_dataset}
\end{table*}

\subsection{Node Classification}

Our experiments are conducted on six benchmark node classification datasets. Tab.~\ref{tab:exp_dataset} presents a statistical summary of these datasets, including their graph properties, domain, and the type of node text.
For consistency, all graphs are treated as
undirected.
We elaborate on each dataset below.

\begin{itemize}[leftmargin=*]
    \item \textbf{Cora~\citep{mccallum2000automating}.}
    In this dataset, nodes represent scientific publications in the machine learning domain, and edges represent citation links between them. The task is to classify each publication into one of seven predefined subject categories (e.g., Neural Networks, Reinforcement Learning). We further divided these nodes into training, validation, and test subsets, using a 7:1:2 ratio.
    
    \item \textbf{Citeseer~\citep{giles1998citeseer}.}
    The Citeseer dataset is another widely-used citation network. Similar to Cora, nodes represent scientific publications and edges represent citations. The classification task is to assign each publication to one of six categories (e.g., AI, DB, IR).
    We further divided these nodes into training, validation, and test subsets, using a 7:1:2 ratio.
    
    \item \textbf{PubMed~\citep{sen2008collective}.}
    The PubMed dataset is a larger citation network sourced from the PubMed medical database. Nodes represent scientific papers, and edges represent citation links. The task is to classify each paper into one of three types. We further divided these nodes into training, validation, and test subsets, using a 7:1:2 ratio.
    To improve computational efficiency for models, we further subsampled the training and validation sets to 10,000 and 300 nodes, respectively.
    The smaller validation set was used for early stopping to prevent overfitting during training.
    
    \item \textbf{History~\citep{yan2023comprehensive}.} In this graph, nodes represent individual books, and an edge connects two nodes if the corresponding books are frequently co-purchased or co-viewed by customers. Node text is the book's title and description. The node classification task is to predict the category of each book from a set of 12 distinct classes. 
    Following the same protocol as with the PubMed dataset, we utilize a subsampled training/validation set.
    
    \item \textbf{Photo~\citep{yan2023comprehensive}.} The Amazon Photo dataset is a co-purchase network from the Amazon-Electronics domain, where nodes represent products and edges indicate frequent co-purchases or co-views. Node text is based on textual reviews, and the task is to classify products into 12 categories.
    We employ the same data partitioning and training set subsampling strategy for this dataset as described for PubMed and History.
    
    \item \textbf{Arxiv~\citep{hu2020open}.} The OGBN-Arxiv dataset is a computer science citation network collected from the Arxiv platform, where nodes represent scientific papers and edges indicate citation links between them. Node text is typically based on the paper title and abstract. The task is to classify each paper into one of 40 predefined subject categories. Based on the original data split, we randomly sample 10,000, 2,000, and 10,000 target nodes for training, validation, and testing, respectively. Crucially, while the supervision signals are downsampled, all applicable models retain access to the complete underlying graph topology for neighbor aggregation, ensuring structural integrity is preserved.
\end{itemize}

\subsection{GraphQA}

We also conduct experiments on the following GraphQA dataset.

\begin{figure*}[!htbp]
    \centering
    \includegraphics[width=1\linewidth]{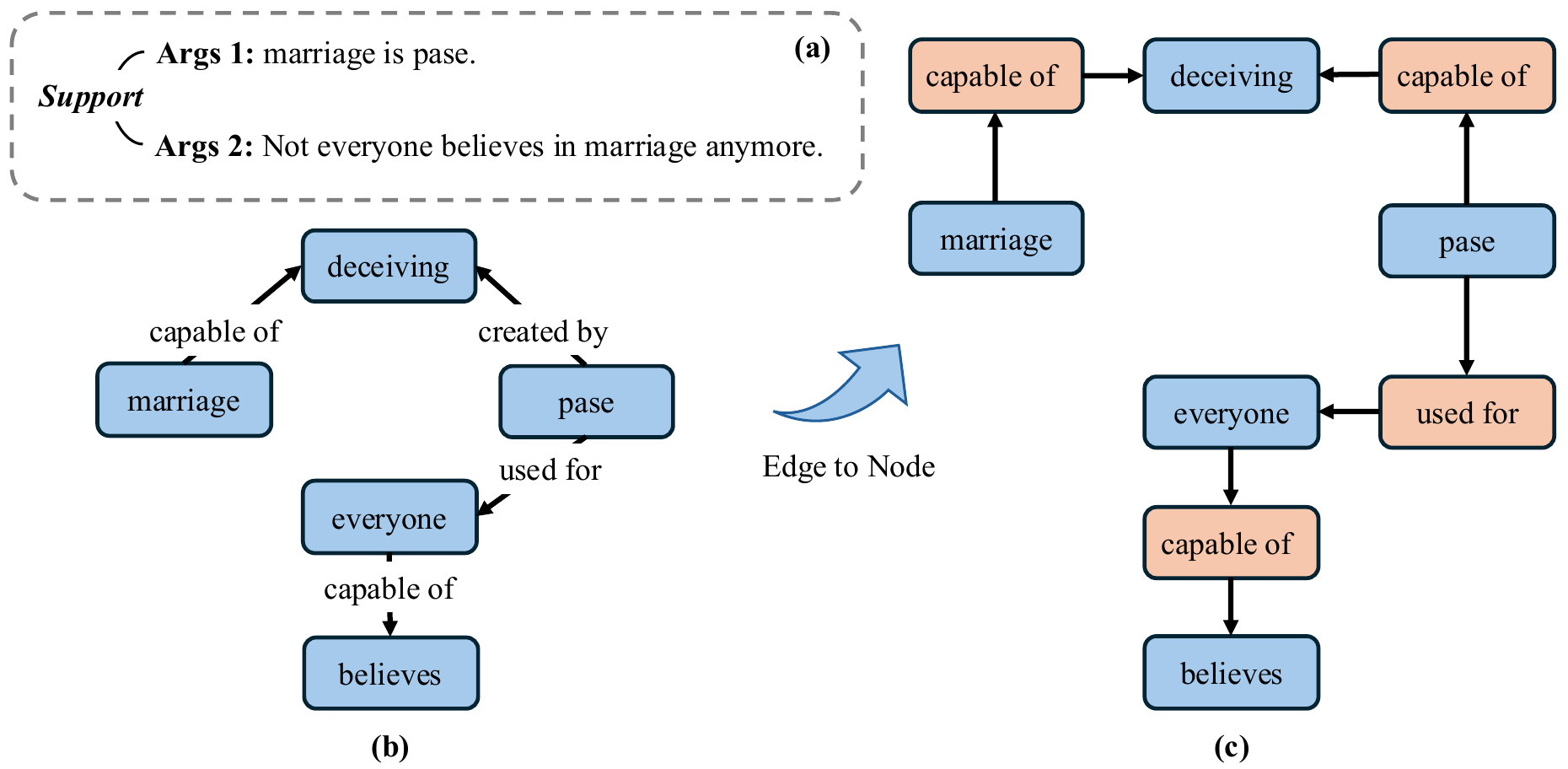}
    \caption{Illustration of the graph transformation process for RAMP in GraphQA task. (a) An example of two arguments with a ``Support'' relation from the ExplaGraphs dataset. (b) The original graph structure where relations are represented as labeled edges. (c) Our transformed graph, where each original edge is reified into a dedicated node (orange), creating a bipartite-like structure that allows RAMP to process textual relational information more effectively.}
    \label{fig:expla_format}
\end{figure*}

\begin{itemize}[leftmargin=*]
    \item \textbf{ExplaGraphs~\citep{saha2021explagraphs}.} This dataset is designed for generative commonsense reasoning.
    It focuses on creating explanation graphs to facilitate stance prediction in debates. The core task is to predict whether arguments are supportive or contradictory to a given belief, with Accuracy serving as the primary metric.
    The graphs in this dataset contain rich semantic information on both nodes and edges. To fully leverage this, we transform the original graphs by converting each edge into a node (as depicted in Fig.~\ref{fig:expla_format}), subsequently creating a bipartite graph structure.
    We follow the data split setting from G-Retriever~\citep{he2024g}, partitioning the dataset into training, validation, and test sets with a ratio of 6:2:2.
\end{itemize}

\section{Baseline}
\label{sec:baselines}

For node classification, we compare RAMP with a range of representative methods widely adopted in prior work. 
In addition, for the GraphQA task, we introduce GNP, G-Retriever, GRAG, and Align-GRAG as the baselines.

\begin{itemize}[leftmargin=*]
    \item \textbf{MLP.} As a non-graph baseline, we use a simple Multi-Layer Perceptron (MLP) that operates solely on node embeddings.
    \item \textbf{GCN~\citep{zhang2019graph}.} We include Graph Convolutional Network (GCN) as a traditional GNN baseline. GCN learns node representations by iteratively aggregating features from their local neighborhoods using a spectral convolution operator.
    \item \textbf{GAT~\citep{velivckovic2017graph}.} A type of GNN with attention weights to differentiate neighbor importance during aggregation.
    This design improves robustness to noisy neighbors, making GAT a representative example of graph models that enhance aggregation through attention mechanisms.
    \item \textbf{GraphSAGE~\citep{hamilton2017inductive}.} As a representative of spatial-domain GNNs, we include GraphSAGE. It learns node embeddings by directly aggregating features from sampled local neighborhoods. This spatial approach provides strong inductive capabilities, allowing the model to generalize to new nodes and graphs.
    \item \textbf{BERT~\citep{devlin2019bert}.} A widely used pre-trained bidirectional transformer encoder. Here, BERT is applied to node texts independently, ignoring graph structure. It serves as an LM-only baseline for evaluating the benefit of structural modeling.
    \item \textbf{RoBERTa~\citep{liu2019roberta}.} We also include RoBERTa, an optimized successor to BERT, to serve as another LM-only baseline.
    RoBERTa enhances the pre-training procedure of BERT through several key modifications: it is trained on a significantly larger corpus, and removes the next-sentence prediction (NSP) objective.
    Similar to the BERT baseline, RoBERTa is directly applied to the node text.
    \item \textbf{Qwen-2.5-7B-Instruct~\citep{qwen2.5}.} A modern decoder-only LLM with strong text understanding and generation capabilities. We adopt its 7B Instruct variant as the backbone in our experiments, using it both as a direct LM-based baseline and as the foundation for building RAMP.
    \item \textbf{GOFA~\citep{kong2024gofa}.} A generative one-for-all model that interleaves GNN-style message-passing layers within an LLM, enabling joint graph and text reasoning. We use its officially released checkpoint for evaluation.
    \item \textbf{OFA~\citep{liu2024one}.} An LLM-GNN hybrid method that unifies diverse cross-domain data and task types by leveraging Text-Attributed Graphs (TAGs) with a single LLM encoder, and introduces a Graph Prompting Paradigm (GPP) based on Nodes-of-Interest (NOI) to enable zero-shot and few-shot classification.
    \item \textbf{LLaGA~\citep{chen2024llaga}.} An LLM-centric method that combines graph with LLM by converting graph into serial text sequences using predefined templates and mapping them into the LLM embedding space with a projector.
    \item \textbf{PromptGFM~\citep{zhu2025llm}.} A recent SOTA ``LLM-as-GNN'' approach that summarizes neighborhoods into language prompts by GPT-4o~\citep{gpt4o2024}, demonstrating the feasibility of LLM-based message passing under context-length constraints. The original implementation relies on GPT-4o~\citep{gpt4o2024}; to enable a fair comparison under comparable computational budgets, we re-implement PromptGFM using the same Qwen-2.5-7B-Instruct backbone as RAMP (denoted \textbf{PromptGFM (Qwen)} in our tables).
    \item \textbf{GNP~\citep{tian2024graph}}. A plug-and-play method that employs a GNN-based module to encode knowledge graphs into prompts for pre-trained LLMs. This prompting module is optimized via a self-supervised link prediction objective.
    \item \textbf{G-Retriever~\citep{he2024g}.} A GraphRAG method for textual graph question answering, which casts subgraph retrieval as a prize-collecting Steiner tree (PCST) to select a compact, query-relevant connected subgraph and generates answers with an LLM. It serves as a representative GraphQA baseline.
    \item \textbf{GRAG~\citep{Hu2024GRAGGR}}. GRAG incorporates textual graphs into LLMs by presenting the context through two complementary text and graph views to enhance the model's comprehension.
    \item \textbf{Align-GRAG~\citep{xu2025align}}. A GraphRAG framework that proposes an aligner to jointly optimize a graph encoder with an LLM-summarized reasoning chain. It employs KL divergence and contrastive loss to simultaneously prune irrelevant knowledge from a retrieved subgraph.
\end{itemize}

To align with the text-rich graph setting, BERT is used for generating node embeddings across all applicable baselines.

\section{Related Work}
\label{sec:related_work}

We organize the related work into three key areas that build the foundation for our proposed RAMP framework: (1) learning on text-rich graphs, which establishes the problem domain; (2) the message passing paradigm in GNNs, which inspires our core aggregation mechanism; and (3) recent efforts to integrate graphs into LLMs, which defines the current state-of-the-art and the specific gap RAMP addresses.

\subsection{Text-Rich Graphs}

A wide range of real-world graphs are text-rich, including knowledge graphs, academic networks, and product graphs~\citep{galkin2023towards,zhang2023effect,zhang2024oag,yang2020large}.
These graphs pose a unique challenge, as models must reason over both complex topological structures and long, nuanced natural language content~\citep{kong2024gofa}.
Early and still-prevalent approaches have been GNN-centric, treating text as a node attribute to be pre-processed~\citep{jin2024large}.
These methods, which we term \textbf{``Compressed Aggregation''} in Fig.~\ref{fig:ramp_intro}(i), first use a text encoder like BERT to collapse each node's text into a single, fixed-size embedding.
This embedding is then propagated through a GNN (e.g., GCN~\citep{zhang2019graph}, GAT~\citep{velivckovic2017graph}).
While computationally efficient, we categorize these as \textbf{``Text-Starved''} learners: the structural propagation occurs entirely in a vector space that is decoupled from the original linguistic evidence.
Even with sophisticated joint training schemes like \textbf{GLEM}~\citep{zhao2022learning} or adapter-based integration as in \textbf{GraphAdapter}~\citep{li2023graphadapter}, the model never ``re-reads'' the text during the structural update, meaning any semantic nuance lost during initialization can never be recovered through graph reasoning. RAMP breaks this paradigm by ensuring that the raw text remains a persistent participant in every message-passing layer.
Another line of research directly pre-trains language models on graph-structured text to enrich node representations~\citep{yasunaga2022linkbert,ye2023language}, but these methods typically focus on enhancing the language model itself rather than defining a native message-passing mechanism for the LLM. 

\subsection{Message Passing in GNNs} Message passing is the central paradigm of most GNNs~\citep{feng2022powerful,papillon2023architectures,zhang2024hierarchical,sun2024towards}, where each node updates its representation by aggregating features from its neighbors.
Foundational models like GCN~\citep{zhang2019graph} and GraphSAGE~\citep{hamilton2017inductive} established the effectiveness of this approach.
A pivotal advancement came with the introduction of attention mechanisms in GNNs, such as GAT~\citep{velivckovic2017graph}, which learn attention coefficients to weight neighbor messages.
This mechanism is particularly insightful for our work.
In an attention-based view, a node’s own representation from the previous layer, $\mathbf{h}_i^{(\ell)}$, serves as the \textbf{query}, while the representations of its neighbors, $\{\mathbf{h}_j^{(\ell)}\}_{j \in \mathcal{N}(i)}$, serve as \textbf{keys} and \textbf{values}.
This asymmetric design is powerful: it allows the target node to \textbf{actively probe} its neighborhood and selectively draw in the most relevant information, rather than being passively overwhelmed by a simple sum or average of all neighbor features.
This prevents the node's accumulated knowledge from being diluted by noisy or irrelevant neighbors, a critical feature for robust representation learning.

\textbf{RAMP's architecture is directly inspired by this principle.} We argue that for a text-rich node, its most potent ``query'' is its \textbf{full, raw text ($\mathbf{X}_i$)}, as it contains the richest semantic context to determine what information is needed from its surroundings. Correspondingly, the \textbf{compact summary tokens of its neighbors ($\mathbf{S}_j^{(\ell)}$)} serve as efficient ``keys'' and ``values'' that can be probed.
By performing attention over its own raw text and its neighbors' summary tokens, RAMP extends this general principle to the LLM setting.
It performs a focused, context-aware aggregation that extracts maximal signal from the neighborhood without the computational burden of processing the full text of every neighbor simultaneously.
\textbf{This design choice is not arbitrary; it is an adaptation of a state-of-the-art GNN principle to the LLM era, positioning RAMP as a natural and powerful evolution for learning on text-rich graphs.}

\subsection{Graphs in LLMs} The rise of LLMs has spurred new paradigms for graph learning~\citep{li2024graph,jin2024large,wang2025graph,guan2024langtopo}.
The most direct approach is \textbf{graph serialization}~\citep{mavromatis2024gnn}, where graph structures and attributes are converted into a linear text sequence for an LLM to process.
This preserves text fidelity but sacrifices \textbf{structural nativity}, forcing the model to infer topological relationships from an unnatural, sequential format and struggling with scalability when node texts are long.
To move beyond naive serialization, an emerging line of work attempts to make LLMs function \emph{as} GNNs, effectively replacing the GNN's aggregation function with an LLM. \textbf{PromptGFM}~\citep{zhu2025llm} uses prompts to iteratively summarize neighborhoods and generate ``language-based IDs'' for nodes. \textbf{LLaGA}~\citep{chen2024llaga} and \textbf{InstructGLM}~\citep{ye2023language} use templates to format local subgraphs into sequences. \textbf{GOFA}~\citep{kong2024gofa}, \textbf{GraphFormers}~\citep{yang2021graphformers}, and \textbf{GL-fusion}~\citep{yang2024gl} inject external, trainable GNN-like layers or adapters into an LLM backbone.
\textbf{OFA}~\citep{liu2024one} employs a different hybrid approach, utilizing LLM to align various graph tasks into a unified format for pre-training general GNNs.
\textbf{HiCom}~\citep{zhang2024hierarchical} explicitly focuses on creating a hierarchical compression scheme to manage long text in large neighborhoods.
While these methods are innovative, they share a common limitation that motivates our work: \textbf{these methods treat the LLM as an external module or a prompted reasoner, rather than recasting it as the fundamental graph operator itself.}
To remain computationally feasible, they all perform some form of information compression or abstraction \emph{before or during} the message aggregation step.
The LLM aggregator in these models never reasons over the explicit, raw text of its multi-hop neighbors.
For example, PromptGFM~\citep{zhu2025llm} reasons on summarized IDs, and HiCom~\citep{zhang2024hierarchical} on compressed tokens. This prevents the LLM from leveraging its greatest strength—deep contextual reasoning on raw evidence—at the most critical step.
This is precisely the gap RAMP fills. By operationalizing the LLM decoder as the core message-passing operator, RAMP goes beyond simulation or external integration.
Unlike serialization, it maintains explicit graph topology via parallel decoding. Unlike other LLM-as-GNN simulations, it maintains a persistent link to the raw text of the target node at every layer. By treating the raw text $\mathbf{X}_i$ as a ``Semantic Anchor'' and the neighborhood $\mathbf{S}_j$ as ``Structural Memory'', RAMP provides an effective framework to achieve deep semantic fidelity and structural integrity simultaneously by redefining the role of the LLM from a component to the core computational engine of the graph learning process.
\section{Additional Experiments}
\label{sec:additional_exps}

\subsection{Impact of Compression Ratio}
\label{ssec:exp_compression_ratio}

To provide principled guidance on selecting the compression ratio $\rho$, we evaluate reconstruction perplexity across different ratios on Cora with $mp\!=\!1$. We measure the same metrics as in Sec.~\ref{ssec:ppl_exp}: Perplexity$_\textit{self}$, which quantifies how faithfully a node's own summary tokens preserve its text (internal fidelity), and Perplexity$_\textit{nbr}$, which measures how well a node's text can be reconstructed from its neighbors' summary tokens (message communicability).

\begin{table}[!htbp]
\centering
\caption{Impact of compression ratio $\rho$ on reconstruction perplexity (Cora, $mp\!=\!1$).}
\small
\begin{tabular}{c|cc}
\toprule
$\rho$ & Perplexity$_\textit{self}$ $\downarrow$ & Perplexity$_\textit{nbr}$ $\downarrow$ \\
\midrule
0.01 & 16.39 & 21.19 \\
0.02 & 13.66 & 21.14 \\
0.05 & 11.36 & 20.68 \\
0.10 & 9.18 & 20.88 \\
\bottomrule
\end{tabular}
\label{tab:exp_compression_ratio}
\end{table}

As shown in Tab.~\ref{tab:exp_compression_ratio}, Perplexity$_\textit{self}$ consistently improves as $\rho$ increases, since more summary tokens better preserve the original text. However, Perplexity$_\textit{nbr}$ saturates quickly: gains in communicability become negligible after $\rho \geq 0.05$, suggesting that a neighbor can only glean a bounded amount of information regardless of summary length. This reveals a ``knee point'' around $\rho \approx 0.05$, where Perplexity$_\textit{self}$ gains begin to diminish while Perplexity$_\textit{nbr}$ has plateaued. We recommend practitioners select $\rho$ at this knee point for new datasets by running a small pre-training evaluation sweep, balancing compression fidelity against computational cost.

\subsection{Impact of Message-Passing Rounds}
\label{ssec:exp_mp_rounds}

In RAMP, one message-passing round ($mp=1$) denotes one graph hop, implemented by neighborhood aggregation followed by a full decoder pass.
To examine the impact of message-passing rounds, we conduct corresponding experiments on datasets from two different domains: 
\textbf{Cora} (Computer Science) and \textbf{PubMed} (Biomedical).

\begin{table}[!htbp]
\centering
\caption{Impact of message-passing rounds (Accuracy \% $\uparrow$).}
\small
\begin{tabular}{c|cc}
\toprule
Method & Cora & PubMed \\
\midrule
Qwen-2.5-7B-Instruct & 82.65 & 90.79 \\
\midrule
RAMP$_{mp=1,\rho=0.1}$ & 83.21 & 92.11 \\
RAMP$_{mp=2,\rho=0.1}$ & \textbf{84.87} & \textbf{93.68} \\
\bottomrule
\end{tabular}
\label{tab:exp_ramp_depth_node_classification}
\end{table}

As shown in Tab.~\ref{tab:exp_ramp_depth_node_classification}, simply using Qwen-2.5-7B-Instruct backbone without any message passing achieves 82.65\% on Cora and 90.79\% on PubMed.  
With RAMP, one round of message passing ($mp\!=\!1$) already brings clear improvements (increasing to 83.21\% on Cora and 92.11\% on PubMed).  
When we further extend the message-passing rounds to two rounds ($mp\!=\!2$), the accuracy further improves by 1.66\% on Cora and 1.57\% on PubMed.
These results suggest that RAMP benefits from deeper propagation, a phenomenon similar to traditional GNNs: multiple rounds of message passing allow nodes to integrate richer contextual signals from their multi-hop neighborhoods, leading to consistent gains across domains.


\subsection{Downstream Ablation of Raw-Text Anchoring}
\label{ssec:exp_compact_downstream}

To validate the effect of raw-text anchoring, we further conduct a downstream ablation that compares full RAMP with a compact variant where raw text is only used at initialization.
As shown in Tab.~\ref{tab:exp_compact_downstream}, removing persistent raw-text anchoring degrades accuracy on all three datasets, with clear dataset-dependent magnitude: 84.87\%\,$\rightarrow$\,55.71\% on Cora, 74.83\%\,$\rightarrow$\,68.22\% on Citeseer, and 93.68\%\,$\rightarrow$\,93.43\% on PubMed. This pattern is consistent with data heterogeneity: while Cora/Citeseer show substantial reliance on continuous anchoring, PubMed exhibits a smaller marginal drop, possibly because its longer and richer node texts provide more informative initial summaries even in the compact setting.

\begin{table}[!htbp]
\centering
\caption{Downstream ablation of raw-text anchoring on node classification (Accuracy \% $\uparrow$). RAMP$_{mp=2,compact}$ denotes the variant where raw text is used only at initialization, and subsequent message passing is performed solely through propagated summary tokens.}
\small
\begin{tabular}{c|ccc}
\toprule
Method & Cora & Citeseer & PubMed \\
\midrule
RAMP$_{mp=2,compact}$ & 55.71 & 68.22 & 93.43 \\
RAMP$_{mp=2,\rho=0.1}$ & \textbf{84.87} & \textbf{74.83} & \textbf{93.68} \\
\bottomrule
\end{tabular}
\label{tab:exp_compact_downstream}
\end{table}

Together with Sec.~\ref{ssec:ppl_exp}, these results suggest that maintaining raw-text access during propagation can be important for downstream graph reasoning.

\subsection{Cross-Subgraph Shuffle: A Radical Topology Ablation}
\label{ssec:exp_cross_subgraph_shuffle}

The shuffle experiments in Sec.~\ref{ssec:exp_shuffling} demonstrate RAMP's permutation invariance and structure sensitivity. To further probe the necessity of topology, we design a more radical ablation: we not only shuffle neighbor order within a subgraph, but also \emph{permute entire subgraphs between samples}, so that each node receives neighbor information that is completely irrelevant to its own content.

\begin{table}[!htbp]
\centering
\caption{Impact of radical topology perturbation on Citeseer (Accuracy \%). The \textit{Within-Subgraph Shuffle} here is equivalent to \textit{Cross-Node Shuffle} in Tab.~\ref{tab:exp_ramp_node_shuffle} of the main text.}
\small
\begin{tabular}{l|c}
\toprule
Setting & Citeseer Acc. \\
\midrule
RAMP (Correct Topology) & 74.83 \\
RAMP (Within-Subgraph Shuffle) & 74.41 $\pm$ 0.01 \\
Fine-tuned Backbone (No Topology) & 74.35 \\
RAMP (Cross-Subgraph Shuffle) & 73.11 $\pm$ 0.06 \\
\bottomrule
\end{tabular}
\label{tab:exp_cross_subgraph_shuffle}
\end{table}

The results in Tab.~\ref{tab:exp_cross_subgraph_shuffle} are definitive. When provided with completely meaningless topology (Cross-Subgraph Shuffle), RAMP's accuracy drops to 73.11\%, falling \emph{below} even the text-only backbone (74.35\%). This demonstrates two key points: (1)~the model is actively reasoning over the provided graph structure rather than ignoring it, and (2)~misleading structural signals are actively harmful, confirming that RAMP's performance gains are fundamentally reliant on meaningful graph topology.

\subsection{Scalability Analysis on PubMed}
\label{ssec:scala_on_pubmed}

\begin{figure}[!htbp]
\setlength{\abovecaptionskip}{2pt}
\setlength{\belowcaptionskip}{-3pt}
    \centering
    \setlength{\abovecaptionskip}{2pt}
    \setlength{\belowcaptionskip}{-3pt}
    \includegraphics[width=0.7\linewidth]{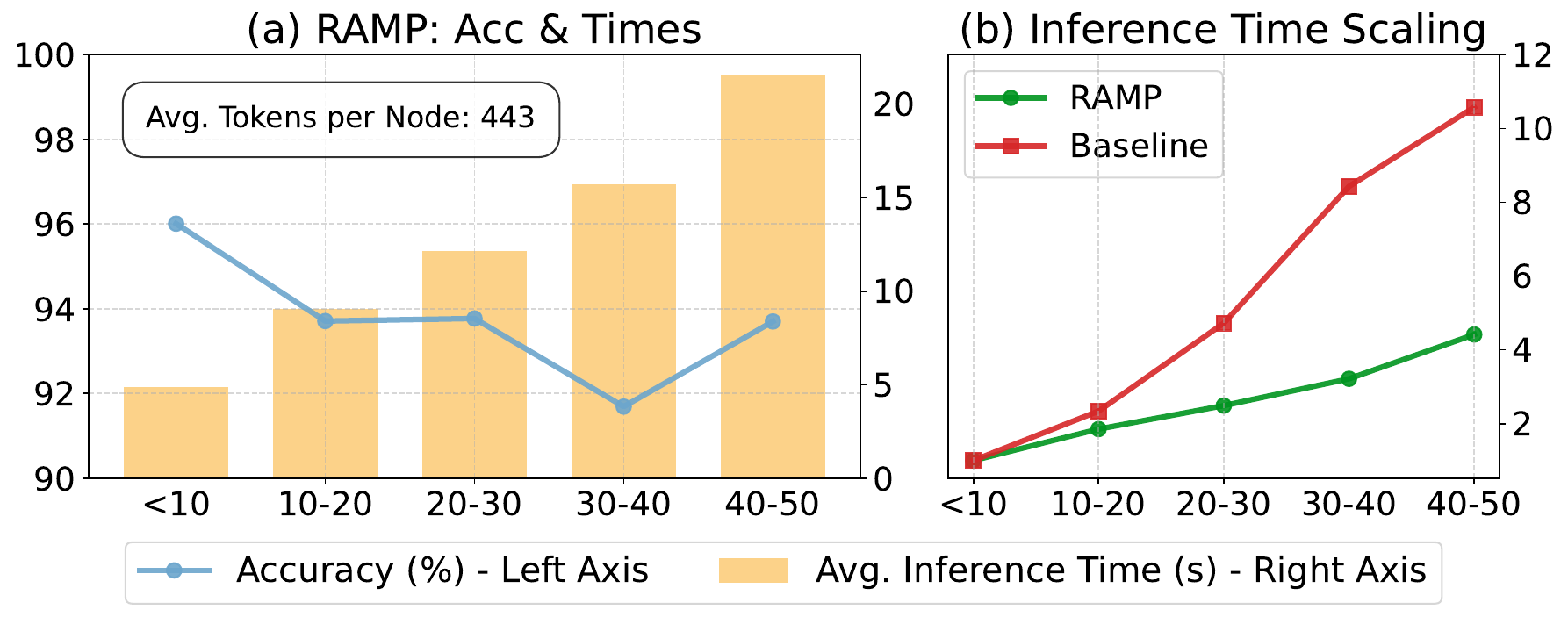}
    \caption{Scalability on PubMed. (a) Accuracy and absolute inference time for RAMP. (b) Inference time scaling for each method, independently normalized to its performance in the smallest bucket (i.e., bucket $<10$).}
    \label{fig:graph_size_scale_pubmed}
\end{figure}

Fig.~\ref{fig:graph_size_scale_pubmed} shows the scalability analysis on PubMed. Despite significantly longer node descriptions (avg. 443 tokens per node), RAMP maintains stable accuracy and exhibits a near-linear scaling profile, similar to results on Cora. Meanwhile, the 0.5B Graph-to-Text baseline demonstrates a much steeper latency growth (Fig.~\ref{fig:graph_size_scale_pubmed}(b)). This cross-dataset consistency confirms that by enabling raw-text anchored message passing via compact summaries, RAMP effectively overcomes the quadratic complexity of traditional Graph-to-Text approaches, even when processing long-text nodes.

\section{Prompt Template Used in RAMP}
\label{sec:prompt_template}

\newtcolorbox{prompt}[2][]{
    breakable,
    colback=black!5!white,
    colframe=black!75!black,
    fonttitle=\bfseries,
    title=#2,
    #1 
}

We provide the prompt templates used in different tasks for reproducibility. 
Each template is divided into \textit{Node Input} (the text content of a node) and \textit{Query Input} (the task-specific instruction given to the model). 

\subsection{Node Classification}

\paragraph{Node Input.} For node classification task, we add \textit{hard prompt} at both the beginning and the end of the node text, 
further guiding the model to summarize the content into a compact representation.

\paragraph{Query Input.} The classification task is formulated as a multiple-choice question, where the model is asked to predict the research 
category of the given node. 
The full node text (title and abstract, wrapped with hard prompts) is also included, ensuring that the model makes predictions conditioned on the same explicit textual information as other backbones.

\begin{prompt}{Prompt Example for Node (Node Classification Task)}

\textit{Above are papers related to the following paper:} [Paper\_ID] Paper \#446271. [Title] Mapping Bayesian Networks to Boltzmann Machines [Abstract] We study the task of tnding a maximal a posteriori (MAP) instantiation of Bayesian network variables, given a partial value assignment as an initial constraint. This problem is known to be NP-hard, so we concentrate on a stochastic approximation algorithm, simulated annealing. This stochastic algorithm can be realized as a sequential process on the set of Bayesian network variables, where only one variable is allowed to change at a time. Consequently, the method can become impractically slow as the number of variables increases. We present a method for mapping a given Bayesian network to a massively parallel Bolztmann machine neural network architecture, in the sense that instead of using the normal sequential simulated annealing algorithm, we can use a massively parallel stochastic process on the Boltzmann machine architecture. The neural network updating process provably converges to a state which solves a given MAP task.
\textit{Please summarise all the given information.}

\end{prompt}


\begin{prompt}{Prompt Example for Query (Node Classification Task)}
Paper \#446271 has the title `Mapping Bayesian Networks to Boltzmann Machines' and the abstract `We study the task of tnding a maximal a posteriori (MAP) instantiation of Bayesian network variables, given a partial value assignment as an initial constraint. This problem is known to be NP-hard, so we concentrate on a stochastic approximation algorithm, simulated annealing. This stochastic algorithm can be realized as a sequential process on the set of Bayesian network variables, where only one variable is allowed to change at a time. Consequently, the method can become impractically slow as the number of variables increases. We present a method for mapping a given Bayesian network to a massively parallel Bolztmann machine neural network architecture, in the sense that instead of using the normal sequential simulated annealing algorithm, we can use a massively parallel stochastic process on the Boltzmann machine architecture. The neural network updating process provably converges to a state which solves a given MAP task.'

Question: Which category should Paper \#446271 be classified as? You can select one from [`Neural\_Networks', `Case\_Based', `Theory', `Reinforcement\_Learning', `Probabilistic\_Methods', `Rule\_Learning', `Genetic\_Algorithms'].
\end{prompt}

\subsection{Graph Question Answering (GraphQA)}
\paragraph{Node Input.} In the GraphQA task, the node/edge texts are relatively short (as in the ExplaGraphs dataset). 
Therefore, we directly feed the raw node text into the model without additional hard prompts.

\paragraph{Query Input.} The query is expressed in natural QA format, requiring the model to reason over the entire graph.


\begin{prompt}{Prompt Example for Node (GraphQA Task)}

created by

\end{prompt}

\begin{prompt}{Prompt Example for Query (GraphQA Task)}

Argument 1: Safe spaces are a redundant and unnecessary practice. Argument 2: Some people have no support or guidance and need it to be available for them.

Question: Do argument 1 and argument 2 support or counter each other? Answer in one word in the form of `support' or `counter'.

\label{prompt:graphqa_query}
\end{prompt}
\section{Generation across Message-Passing Rounds: A Case Study}
\label{app:generation_across_layer}

To better understand what happens during the propagation process of RAMP, 
we compare model outputs when conditioning on the \textbf{penultimate} message-passing round KV versus the \textbf{last} round KV of message passing. 
We provide representative cases from the \textbf{Cora} dataset in Tab.~\ref{tab:cora_layer_case}, 
which illustrate how intermediate and final outputs can differ. 
Based on these observations, several consistent patterns emerge:

\begin{itemize}[leftmargin=*]
    \item Even though only the last round is trained to match the ground truth, 
    the penultimate round already produces \textbf{non-trivial generations} that often resemble intermediate analysis. 
    
    \item When the penultimate round prediction is incorrect, the final round sometimes \textbf{corrects it}, likely due to having aggregated information from a larger neighborhood scope.
    
    \item Conversely, we also observe cases where the final round prediction is \textbf{misled by} an erroneous intermediate generation. 
    
    \item Overall, these qualitative results suggest that intermediate rounds are not merely placeholders for information aggregation, sometimes exhibiting reasoning-like traces before the final prediction.
\end{itemize}

\paragraph{Connection to Multi-Agent Communication.} These qualitative patterns support viewing RAMP as a multi-agent communication framework~\citep{zhuge2024gptswarm, qian2024scaling, zou2025latent}. In this view, each decoder acts as an individual agent, and message-passing serves as a communication protocol. The ``reasoning-like traces'' observed in intermediate rounds represent an iterative refinement process, where agents exchange information via summary tokens to resolve ambiguities and reach a final consensus.

\begin{table*}[!th]
\scriptsize
\renewcommand\arraystretch{1.2}
\setlength{\tabcolsep}{2pt}
\caption{Examples showing generations from different message-passing rounds in RAMP. 
While only the final round is supervised during training, we observe that the penultimate round KV also produces meaningful outputs. 
Comparisons across rounds reveal intermediate reasoning behaviors, occasional error correction, and in some cases error propagation. 
\textit{Italic text} indicates predictions that do not match the ground truth. }

\begin{tabularx}{\textwidth}{>{\raggedright\arraybackslash}m{0.20\textwidth} 
>{\raggedright\arraybackslash}m{0.30\textwidth} 
>{\raggedright\arraybackslash}m{0.25\textwidth}
>{\raggedright\arraybackslash}m{0.25\textwidth} }
\toprule
\textbf{Paper Title} & \textbf{\makecell{Intermediate Output \\ (penultimate round KV)}} & \textbf{\makecell{Final Output \\ (last round KV)}} & \textbf{Ground Truth}   \\ \midrule

Unsupervised Learning Using MML & This paper should be classified into: \textit{Theory methods} & This paper should be classified into: Probabilistic\_Methods & This paper should be classified into: Probabilistic\_Methods  \\  \midrule

First Order Regression  &  This paper should be classified into: \textit{Theory}  & This paper should be classified into: Rule\_Learning & This paper should be classified into: Rule\_Learning \\  \midrule

First Order Regression: Applications in Real-World Domains  & This paper should be in the rule-based category as it deals with rule based machine learning  &  This paper should be classified into: Rule\_Learning & This paper should be classified into: Rule\_Learning \\  \midrule

Probably Approximately Optimal Derivation Strategies  &  This paper should be classified into: Theorys category: theory abstract: an paper describes a pao algorithm that first uses mu set of training examples to approximate these probability values and then uses these estimates to produce mu ``probably approximately optimal'' strategy that is given any *; ffi > 0, pao produces mu strategy whose cost is within * of mu cost of mu optimal strategy with probability greater ... &  This paper should be classified into: Theory  & This paper should be classified into: Theory \\ \midrule


A Decision Tree System for Finding Genes in DNA  &  This paper should be in \textit{theory} category, it describes a decision tree system for finding genes in vertebrate dna sequences, and uses new methods for identifying start codons, donor sites, and acceptor sites, and brings them together in a frame sensitive dynamic programming algorithm that findss the minimum segmentation of a dna sequence into coding and ...  & This paper should be classified into: Neural\_Networks  & This paper should be classified into: Neural\_Networks \\  \midrule


Space-Frequency Localized Basis Function Networks for Nonlinear System Estimation and Control  &  This paper should be classified into: Neural This paper deals with the on-line generation of nonparametric models for nonlinear systems using tools from multiresolution analysis and wavelet theory. It extendss earlier results on adaptive control and identification of nonparametric systems using wavelet basis functions to on-line generation of nonparametric models for nonparametric systems with wavelet basis functions ... &  This paper should be classified into: Neural\_Networks & This paper should be classified into: Neural\_Networks \\  \midrule


Gas Identification System using Graded Temperature Sensor and Neural Net Interpretation  & This paper should be classified into: Neural\_ interpretation (\textit{or} case-based reasoninging) as it uses a (fuzzy logic) for interpretation of sensor data.  &  This paper should be classified into: Neural\_Networks & This paper should be classified into: Neural\_Networks \\  \midrule

Using Partitioning to Speed Up Specific-to-General Rule Induction  &  This paper should be classified into: \textit{rule induction (theory)} & This paper should be classified into: \textit{Rule\_Learning}  & This paper should be classified into: Case\_Based \\  \midrule

Genetic Algorithm based Scheduling in a Dynamic Manufacturing Environment  &  This paper should be classified into: Genetic Alculms &  This paper should be classified into: Genetic\_Algorithms & This paper should be classified into: Genetic\_Algorithms \\  \midrule

Combining Rules and Cases to Learn Case Adaptation  & This paper should be in the case-based category. It describes a new approach to learn case adaptation knowledge in case-based reasoning (cbb) systems. The paper is not about neural networkss, theory, reinforcement learning, probabilistic methods, rule learning, or genetic algorithmss.  &  This paper should be classified into: Case\_Based & This paper should be classified into: Case\_Based \\   \midrule

Associative Reinforcement Learning: Functions in k-DNF &  This paper should be classified into: Reinmunl Learning & This paper should be classified into: Reinforcement\_Learning & This paper should be classified into: Reinforcement\_Learning \\

\bottomrule
\end{tabularx}
\label{tab:cora_layer_case}
\end{table*}


\end{document}